\documentclass[sigconf,nonacm]{acmart}

\def\BibTeX{{\rm B\kern-.05em{\sc i\kern-.025em b}\kern-.08emT\kern-.1667em\lower.7ex\hbox{E}\kern-.125emX}}

\usepackage{subcaption}
\usepackage{graphicx}
\usepackage{amsmath}
\usepackage{amsfonts}

\begin{document}

%
\title{Precision-Weighted Federated Learning}

%

\author{Jonatan Reyes}
\email{jonatan.reyes@mail.concordia.ca}
\affiliation{
    \institution{Concordia University}
    \city{Montreal}
    \country{Canada}
}

\author{Lisa Di Jorio}
\email{lisa@imagia.com}
\affiliation{
    \institution{Imagia Cybernetics Inc.}
    \city{Montreal}
    \country{Canada}
}

\author{Cecile Low-Kam}
\email{cecile.low-kam@imagia.com}
\affiliation{
    \institution{Imagia Cybernetics Inc.}
    \city{Montreal}
    \country{Canada}
}

\author{Marta Kersten-Oertel}
\email{marta.kersten@concordia.ca}
\affiliation{
    \institution{Concordia University}
    \city{Montreal}
    \country{Canada}
}

%
\renewcommand{\shortauthors}{Reyes et al.}

%
\begin{abstract}
Federated Learning using the Federated Averaging algorithm has shown great advantages for large-scale applications that rely on collaborative learning, especially when the training data is either unbalanced or inaccessible due to privacy constraints. We hypothesize that Federated Averaging underestimates the full extent of heterogeneity of data when the aggregation is performed. We propose \emph{Precision-weighted Federated Learning}\footnote{A US provisional patent application has been filed for protecting at least one part of the innovation disclosed in this article} a novel algorithm that takes into account the second raw moment (uncentered variance) of the stochastic gradient when computing the weighted average of the parameters of independent models trained in a Federated Learning setting.  With Precision-weighted Federated Learning, we address the communication and statistical challenges for the training of distributed models with private data and provide an alternate averaging scheme that leverages the heterogeneity of the data when it has a large diversity of features in its composition. Our method was evaluated using three standard image classification datasets (MNIST, Fashion-MNIST, and CIFAR) with two different data partitioning strategies (independent and identically distributed (IID), and non-identical and non-independent (non-IID)) to measure the performance and speed of our method in resource-constrained environments, such as mobile and IoT devices. The experimental results demonstrate that we can obtain a good balance between computational efficiency and convergence rates with Precision-weighted Federated Learning. Our performance evaluations show $9\%$ better predictions with MNIST, $18\%$ with Fashion-MNIST, and $5\%$ with CIFAR-10 in the non-IID setting. Further reliability evaluations ratify the stability in our method by reaching a 99\% reliability index with IID partitions and 96\% with non-IID partitions. In addition, we obtained a $20x$ speedup on Fashion-MNIST with only 10 clients and up to $37x$ with 100 clients participating in the aggregation concurrently per communication round. The results indicate that Precision-weighted Federated Learning is an effective and faster alternative approach for aggregating private data, especially in domains where data is highly heterogeneous. 

\end{abstract}

\maketitle

%

%

\section{Introduction}

Machine learning based on distributed deep neural networks (DNNs) has gained significant traction in both research and industry~\cite{lecun2015deep, najafabadi2015deep}, with many applications in IoT, for mobile devices, and in the automobile sector. For example, IoT devices and sensors can be protected from web attacks during the exchanging of data between the device and web services (or data stores) in the cloud \cite{tian2019distributed}. Mobile devices use distributed learning models to assist in vision tasks for automatic corner detection in photographs~\cite{rosten2006machine}, prediction tasks for text entry \cite{yang2018applied}, and recognition tasks for image matching and speech recognition \cite{lecun2015deep}. Alternatively, modern automobiles utilize distributed machine learning models to improve drivers' experience, vehicle's self-diagnostics and reporting capabilities \cite{johanson2014big}. 


Despite the benefits provided by distributed machine learning, data privacy and data aggregation are raising concerns addressed in various resource-constrained domains. For example, the communication costs incurred when updating deep learning models in mobile devices is expensive for most users as their internet bandwidths are typically low. In addition, the data used during the training of models in mobile devices is privacy-sensitive, and operations of raw data outside the portable devices are susceptible to attacks. One solution is using secure protocols \cite{bonawitz2016practical} or differential-privacy guarantees \cite{dwork2014algorithmic, melis2019exploiting} to ensure that data is transferred between clients and servers safely. Another solution is to use data aggregation for distributed DNNs mitigating the need for transferring data to a central data store. With this solution, the learning occurs at the client level where models are optimized locally across the distributed clients. This approach is termed \emph{Federated Learning} \cite{mcmahan2016communication}.

McMahan \emph{et al.} \cite{mcmahan2016communication} introduced the notion of Federated Learning in a distributed setting of mobile devices. Their developed \emph{Federated Averaging} algorithm uses numerous communication rounds where all participating devices send their local learning parameters, i.e. DNN weights, to be aggregated in a central server in order to create a global shared model. Once the global model is computed, it is distributed to every client replacing the current deep learning model. Since only the global model is communicated in these rounds, data aggregation is achieved, even though the client's raw data never leaves the device. Given such a setup, individual clients can collaboratively learn an averaged shared model without compromising confidentiality. This makes Federated Learning a promising solution to the analysis of privacy-sensitive data distributed across multiple clients.

In McMahan \emph{et al.'s} original paper the local learning parameters on each client are aggregated by the central server and the global model is maintained with the \emph{weighted average} of these parameters \cite{mcmahan2016communication}. There are potentially a few statistical shortcomings identified with this type of averaging method. If we consider that the aggregation of weights across multiple clients is similar to a meta-analysis which synthesizes the effects of diversity across multiple studies then variation across the population should be considered. Meta-analysis is a quantitative method that combines results from different studies on the same topic in order to draw a general conclusion and to evaluate the consistency among study findings\cite{hedges_olkin_1985, nakagawa2012methodological}. There is compelling evidence that demonstrates a misleading interpretation of results and a reduction of statistical power when combining data from different sources without accounting for variation across the sources \cite{ioannidis2007heterogeneity, bangdiwala2016statistical, lin2010meta}.  

\subsection{Hypotheses}

\begin{figure}
  \centering
  \includegraphics[width=50mm]{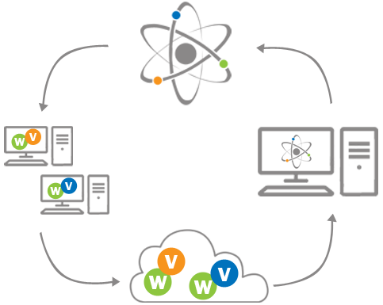}
  \caption{Aggregation of weights and variance via Precision-weighted Federated Learning: local models are trained across clients \emph{(Left)}, weights and variances are aggregated by the central server \emph{(Bottom)}, a centralized model is computed \emph{(Right)} and the aggregated weights are redistributed across clients \emph{(Top)}.}
  \Description{Precision-weighted Federated Learning}
  \label{fig:aggregation_weights}
\end{figure}

In this paper, we build on the work of McMahan \emph{et al.}\cite{mcmahan2016communication}, and propose the \emph{Precision-weighted Federated Learning} algorithm, a novel \emph{variance-based} averaging scheme to aggregate model weights across clients. The proposed method penalizes the model uncertainty at the client level to improve the robustness of the centralized model, regardless of the data distribution: independent and identically distributed (IID) or non-identical and non-independent (non-IID). Our approach makes use of the uncentered variance of the gradient estimator from the Adam optimizer \cite{kingma2014adam} to compute the weighted average at each communication step (Figure \ref{fig:aggregation_weights}). 

We hypothesize that the Federated Averaging algorithm underestimates the full extent of heterogeneity on domains where data is complex with a large diversity of features in its composition. More specifically, we hypothesized that: (1) Precision-weighted Federated Learning can leverage individual intra-variability when averaging multiple sources to improve performances when the training data is highly-heterogeneous across sources, and (2) it can harness individual intra-variability when averaging multiple sources to accelerate the learning process, especially when data is highly-heterogeneous across sources. 

To test our hypothesis we compared the performance of the original Federated Averaging algorithm against the Precision-weighted method in a number of image classification tasks using MNIST \cite{lecun1998gradient}, Fashion-MNIST \cite{xiao2017fashion}, and CIFAR-10 \cite{krizhevsky2009learning} datasets.

\subsection{Contributions}

The contributions of this paper are threefold: (1) We propose a novel algorithm for the averaging of distributed models using the estimated variance of the stochastic gradient computed independently by each client in a Federated Learning environment; (2) We provide extensive evaluations of the method using benchmark image classifications datasets demonstrating its robustness to unbalanced and non-IID data distributions; and (3) We compare the method to Federated Averaging on empirical experiments, and with fewer communication rounds we obtain comparable accuracy on IID distributions, greater accuracy on non-IID distributions, and more stable accuracy over communication rounds over all distributions.

\section{Related Work}

There is an increasing concern for aggregation of private data in the data mining domain, particularly when models require access to a client's data in order to improve their accuracy \cite{agrawal2000privacy, lindell2000privacy}. Data privacy and data aggregation are thus concerns that are actively being investigated for both centralized \cite{xu2014information, verykios2004state} and decentralized (or distributed) \cite{mcmahan2016communication} data environments.

The method proposed in this paper is dedicated to the aggregation of weights for DNNs with decentralized data. It is, therefore, important to observe the communication challenges addressed in previous work, mainly the security and protection of data, and the reduction of the steps needed in communication cycles. Bonawitz \emph{et al.} \cite{bonawitz2016practical} proposed a complementary approach to Federated Learning: a communication-efficient secure aggregation protocol for high-dimensional data. In Bonawitz \emph{et al.}'s work, Federated Learning was used in the training of DNN models for mobile devices using secure aggregation algorithms to protect the data residing on individual mobile devices. On the other hand, Kone\u cn\`y \emph{et al.} \cite{konevcny2016federated} presented two optimization algorithms (structural and sketched updates) to reduce the communication cost in the training of deep neural networks in a federation of participant mobile devices. 

As well as communication challenges, the aggregation of data in decentralized environments is impacted by statistical challenges, especially when the training data is non-IID. Smith \emph{et al.} \cite{smith2017federated} highlighted the fact that data across a DNN is often non-IID distributed; that is, each participant updating the shared model in a Federated Learning setting generates a distinct distribution of data. One way to handle data heterogeneity is by using multi-task learning (MTL) frameworks. Smith \emph{et al.} created the MOCHA framework, which enables the analysis of data variability in a Federated MTL. However, as noted by Zhao \emph{et al.} \cite{zhao2018federated}, the Federated MTL is not comparable with the original work on Federated Learning as the proposed framework does not apply to non-convex deep learning models. In the same paper, Zhao \emph{et al.} proposed a data-sharing strategy to improve test-accuracy when data is non-IID. This method requires a small subset of data consisting of a uniform distribution to be shared across clients. Albeit promising results can be achieved with this method, the shared subset of data may not always be available, especially when data is highly sensitive in nature. Other methods explore the statistical challenges of Federated Learning by creating synthetic data using Dirichlet distributions with different concentration parameters. This technique allows the creation of more realistic non-IID data distributions at the client level, which are used to examine the effects on aggregations carried out with the Federated Learning algorithm \cite{zhao2018federated, hsu2019measuring}.

There is a diverse body of work that further explores collaborative learning, data sharing and data preservation across multiple data centers. Note that all of these methods are substantially different than the original work on Federated Learning. Although some yield comparable or even better results than Federated Learning they lack empirical observations with non-IID data. Chang \emph{et al.} \cite{chang2018distributed} addressed the problem of distributed learning on medical data and compared five heuristics: separate training on subsets, training on pooled data, weight averaging, and weight transfer (single and cyclical transfer). Of all these heuristics, training on pooled data has the best prediction performance and training on cyclical weight transfer achieved comparable testing accuracy to that of centrally trained models. Xu \emph{et al.} \cite{xu2018collaborative} introduced a collaborative deep learning (co-learning) method for the training of a shared global model using a cyclical learning rate schedule mixed with an incremental number of epochs. Their results demonstrate that the method is comparable with data centralized learning. Lalitha \emph{et al.} \cite{lalitha2018fullydecentralized} trained a model over a network of devices without a centralized controller. However, the users could communicate locally with their closest neighbors. The performance of the proposed algorithm on two users matches the performance of an algorithm trained by a central user with access to all data. They left a full empirical evaluation for future research. Chen \emph{et al.} \cite{chen2018federated} proposed a Federated Meta-Learning framework for the training of recommended systems. The framework permits data sharing at the algorithm level, preserves data privacy, and reports an increase of 12.4\% in accuracy compared with previous results. Kim \emph{et al.} \cite{kim2018keep} addressed the problem of catastrophic forgetting (the ability of neural networks to learn new tasks while discarding knowledge about previous learned tasks) in a distributed learning environment on clinical data and introduced an approach for knowledge preservation. Similarly, Bui \emph{et al.} \cite{bui2018partitioned} unify continual learning and Federated Learning in a partitioned variational inference framework. Vepakomma \emph{et al.} \cite{vepakomma2018splitlearning}, introduced \textit{split learning}, which addresses challenges specific to health data, such as different modalities across clients, no label sharing and semi-supervised learning.

In the field of genetics, genome-wide association studies aim to identify genetic variants associated to phenotypes of interest. As the effect of these variants on phenotypes is usually moderate, individual hospital studies are under-powered to detect them with confidence and a growing number of consortia are created to combine data across studies. As patient genotypes are privacy sensitive, these consortia use \textit{meta-analyses} to aggregate summary statistics from multiple studies. This increases the statistical power of finding a mutation related to a phenotype, while protecting the privacy of individual genotypes. Lin and Zang \cite{lin2010meta} demonstrated that meta-analyses achieve comparable efficiency as analyses of pooled individual participants under mild assumptions. This proximity with the distributed learning setting motivated us to create the Precision-weighted Federated Learning, an averaging approach that considers a meta-analysis weighting scheme in the aggregation of the effects of the variances from the weights generated during training of the neural network.

\section{PRECISION-WEIGHTED FEDERATED LEARNING}

The Precision-weighted Federated Learning approach combines the weights from each client into a globally shared model where the aggregation is achieved by averaging the weights by the inverse of their estimated variance. We will use the same notations than the Federated Averaging algorithm \cite{mcmahan2016communication} to describe the implementation of the proposed method. We consider the general objective
\begin{eqnarray}
\min_{w \in \mathbb{R}^d} f(w) & \text{with} & f(w) = \frac{1}{n} f_i(w),
\label{eq:objective}
\end{eqnarray}
for $i=1, ..., n$, where $n$ is the number of data examples and $f_i(w) = \ell(x_i, y_i; w)$ is the loss of the prediction on example $(x_i, y_i)$ made with model parameters $w$. If the data is partitioned over $K$ clients, McMahan \emph{et al.} rewrite the objective of Equation \ref{eq:objective} as
\begin{eqnarray}
f(w) = \sum_{k=1}^K \frac{n_k}{n} F_k(w) & \text{with} & F_k(w) = \frac{1}{n_k} \sum_{i \in \mathcal{P}_k} f_i(w),
\end{eqnarray}
where $\mathcal{P}_k$ is the set of indexes of data examples on client $k$ and $n_k = |\mathcal{P}_k |$.
Under a uniform distribution of training examples over the clients, the \emph{IID assumption}, the expectation of the client-specific loss $F_k(w)$ is $f(w)$. In a non-IID setting however, this result does not hold \cite{mcmahan2016communication}.

The corresponding stochastic gradient descent for optimization with a fixed learning rate $\eta$ consists in computing the gradient
\begin{eqnarray}
g_k & = & \nabla F_k(w_t) 
\end{eqnarray}
for each client $k$ at iteration $t$, and applying the two successive updates
\begin{eqnarray}
w_{t+1}^k & \leftarrow & w_t - \eta g_k \text{ for all } k
\end{eqnarray}
\begin{eqnarray}
w_{t+1} & \leftarrow & \sum_{k=1}^K \frac{n_k}{n} w_{t+1}^k.
\label{eq:global_update}
\end{eqnarray}

With Precision-weighted Federated Learning the global update of Equation \ref{eq:global_update} is replaced by
\begin{eqnarray}
w_{t+1} \leftarrow \sum_{k=1}^K \frac{\left(v^k_{t+1}\right)^{-1}}{\sum_{k=1}^K \left(v^k_{t+1}\right)^{-1}} w_{t+1}^k
\label{eq:pw}
\end{eqnarray}
where $v^k_{t+1}$ denotes the variance of the maximum likelihood estimator of weight $w$ at iteration $t+1$ for client $k$. This inverse variance weighting scheme used in Equation \ref{eq:pw} corresponds to the fixed effect model used in meta-analyses. Intuitively, this method allows taking into consideration the uncertainty of each client into the aggregated result and uses the estimated variance to penalize the model uncertainty at the client level: models with high estimated variance across clients have a smaller impact on the aggregation result at the current communication round. Although $v^k_{t+1}$ is inversely proportional to the sample size, it is a more nuanced summary as it captures additional uncertainty about the client's weights.

To estimate the inverse of the variance of the maximum likelihood, we use the raw second moment estimate (uncentered variance) from the Adam optimizer \cite{kingma2014adam}, which approximates the diagonal of the Fisher information matrix \cite{pascanu2014natural}. Our experiments show that this approximation manages to capture the uncertainty of weights in practice.

\section{METHODOLOGY}\label{sec:methodology}

We tested the Precision-weighted Federated Learning method under different data distributions for image classification tasks. The baseline we use is the Federated Averaging approach. Firstly, we explore the performance of our method in resource-constrained environments, applicable to areas where memory is limited, such as mobile and IoT devices. Next, we present a scenario in which we investigate the speedup of our method as a function of the number of clients participating in the aggregation of weights. Finally, we present the analysis for the generalization of the global model when the parameter variance is applied to the aggregation of parameters of all the models in the distributed learning process.

Since the statistics of the data are influenced by the way it is distributed across clients, we tested the proposed methodology with both IID and non-IID data distributions. To create these scenarios, we distributed the training data across individual clients in two configurations (see Section \ref{ss:4.2}). The complexity of the image recognition problems was increased in agreement with the methodology proposed by Scheidegge \emph{et al.} \cite{scheidegger2018efficient} and therefore MNIST, Fashion-MNIST and CIFAR-10 were used as benchmarks. Furthermore, we utilized modest convolutional architectures to
compare training speed and optimal convergence with our method and Federated Averaging and to explain the effects of variance in the generalization of the centralized model. All of the experiments were executed on an NVIDIA Tesla V100 Graphic Processing Unit. 

\subsection{Datasets}
\noindent \textbf{MNIST}: The MNIST dataset consist of 70,000 gray-scale images (28 x 28 pixels in size) which are divided in 60,000 training and 10,000 test samples. The images are grouped in 10 classes corresponding to the handwritten numbers from zero to nine. 

\noindent \textbf{CIFAR-10}: The CIFAR-10 dataset consists of 60,000 colored images (36 x 36 pixels in size) divided in a training set of 50,000 and a testing set of 10,000 images. Images in CIFAR-10 are grouped into 10 mutually exclusive classes of animals and vehicles: airplanes, automobiles, birds, cats, deer, dogs, frogs, horses, ships, and trucks.

\noindent \textbf{Fashion-MNIST}: The Fashion-MNIST dataset contains the same number of samples, image dimensions and number of classes (different labels) in its training and testing sets than MNIST, however, the images are of clothing (e.g. t-shirts, coats, dresses and sandals). 

\subsection{Data Distributions}\label{ss:4.2}

\begin{figure}[h!]
\begin{subfigure}{0.33\textwidth}
\includegraphics[width=1.1\linewidth, height=4cm]{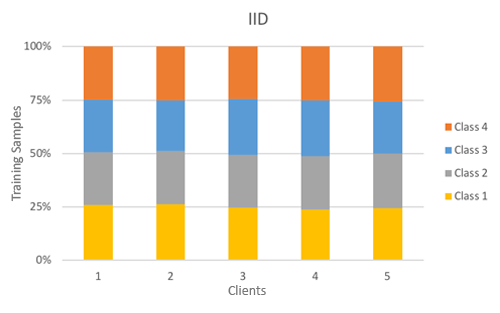} 
\caption{IID}
\label{fig:subim1}
\end{subfigure}
\hspace{1cm}
\begin{subfigure}{0.33\textwidth}
\includegraphics[width=1.1\linewidth, height=4cm]{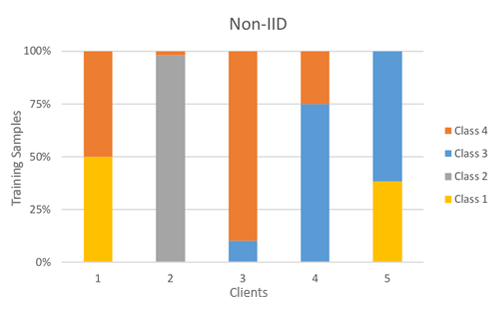}
\caption{non-IID}
\label{fig:subim2}
\end{subfigure}

\caption{Example of data distributions for 4 classes and 5 clients.}
\Description{data distributions}
\label{fig:data_distribution}
\end{figure}

\noindent \textbf{IID}: With IID data distribution the number of classes and the number of samples per class were assigned to clients with a uniform distribution. We shuffled the training data and created one partition per client with an equal number of samples per class. For example, 10 clients receive 600 samples per class. Figure \ref{fig:data_distribution} \textit{(Top)} shows an example with 5 clients and 4 classes.

\noindent \textbf{Non-IID}: With this data partition, two classes are assigned per client at most. This is similar to the partition shown in \cite{mcmahan2016communication} used to explore the limits of the Federated Averaging approach, which we now use to test and compare our algorithm under similar circumstances. In this extreme scenario, the number of samples per class per client is evenly distributed, creating a balanced scenario. (Figure \ref{fig:data_distribution} \textit{(Bottom)}).

\subsection{Convolutional Neural Networks}
The architectures used in our experiments were CNNs trained from scratch. All artificial networks were based on the Keras sequential model, trained with the Adam optimizer and an objective function as defined by categorical cross-entropy.

For MNIST and Fashion-MNIST the architecture of the first artificial neural network consisted of two convolutional layers using 3x3 kernels (each with 32 convolution filters). A  rectified  linear  unit (ReLU) activation is performed right after each convolution, followed with a 2x2 max pooling used to reduce the spatial dimension, a dropout layer used to prevent overfitting, a fully densely-connected layer (with 128 units using a ReLu activation), and leading to a final softmax output layer (600,810 total parameters). The network was trained from scratch using partitions of training data and the final model was evaluated using the testing set. 

A second network was used to train our models using data from the CIFAR-10 dataset. The architecture consisted of one 3x3 convolutional layer (with 32 convolution filter using a ReLu activation), followed with a 2x2 max pooling, a batch normalization layer; a second 3x3 convolutional layer (with 64 convolution filter using a ReLu activation), followed with, a batch normalization layer and a 2x2 max pooling; a dropout layer; one fully densely-connected layer (with 1024 and 512 units using a ReLu activation), another dropout layer; and a final softmax output layer (4,225,354 total parameters).

\subsection{Adam and the Weighted-Variance Callback}

A key component in the formulation of the weighted average algorithm is the estimation of the individual intra-variability expressed during the training of local data. As the training of the model proceeds, we capture the weights' variances via the second raw moment (uncentered variance) of the stochastic gradient descent from the Adam optimizer and use it in the construction of the Precision-weighted Federated Learning algorithm. In order to access the internal statistics of the model during training, we use a callback function that averages the variance estimators on the second half of the last epoch. The last epoch is chosen as it provides a more accurate prediction of the variance of the final weight.

\section{RESULTS}

This section presents the results of our model predictions trained with the two aforementioned data partitioning strategies in Section \ref{sec:methodology} and demonstrates the limits and practical application of the proposed method. All of our experiments use a different random seeds to randomize the order of observations during the training of the local models.  As noted in McMahan \emph{et al.'s} paper, averaging federated models from different initial conditions leads to poor results. Thus, in order to avoid the drastic loss of accuracy observed on independent initialization of models for general non-convex objectives, each local model was trained using a shared random initialization for the first round of communication. After the first round of communication, all local models were initialized with the globally averaged model aggregated from the previous round. 

\subsection{Evaluating Computational Resources}

Experiments with MNIST and Fashion-MNIST were conducted by using 500 rounds of communication, 1 epoch, and batch sizes (10, 25, 50, 100, and 200). Similarly, experiments with CIFAR-10 were executed for 500 rounds of communication, with 10 epochs, and batch sizes (10, 25, 50, 100, and 200). All of the training samples of each dataset were arranged among 10 clients.

\begin{figure}[h!]
\begin{subfigure}{0.33\textwidth}
\includegraphics[width=1.1\linewidth, height=4cm]{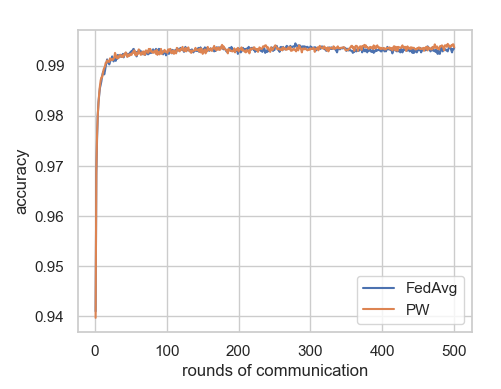} 
\caption{MNIST (B = 50)}
\label{fig:subim1}
\end{subfigure}
\begin{subfigure}{0.33\textwidth}
\includegraphics[width=1.1\linewidth, height=4cm]{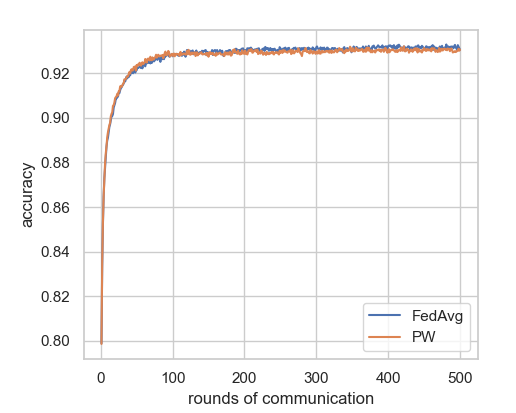}
\caption{Fashion-MNIST (B = 50)}
\label{fig:subim2}
\end{subfigure}
\begin{subfigure}{0.33\textwidth}
\includegraphics[width=1.1\linewidth, height=4cm]{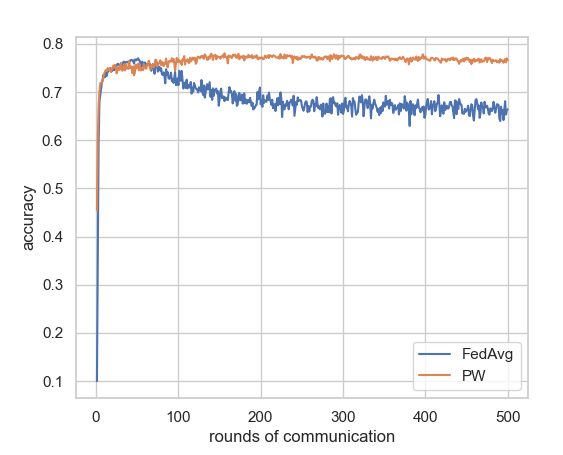}
\caption{CIFAR-10 (B = 10)}
\end{subfigure}
\caption{Test-accuracy for Federated Averaging (FedAvg) and Precision-weighted Federated Learning (PW) using IID data distributions.}
\Description{MNIST, Fashion-MNIST, CIFAR with IID data distributions}
\label{fig:iid_mnist_fashion}
\end{figure}

The comparison results of test-accuracy between Federated Averaging and Precision-weighted Federated Learning aggregation methods using IID partitions is given in Table ~\ref{tab:iid}.  Given this setup, test-accuracy scores are comparable with those obtained using Federated Averaging, however, our method is more stable. When we analyze the results of MNIST and Fashion-MNIST, we observe that test-accuracy values are consistent across batch sizes. The accuracy curves of Precision-weighted Federated Learning and the Federated Averaging for these datasets are show in Figure \ref{fig:iid_mnist_fashion} \emph{(a)} and \emph{(b)}). Alternatively, CIFAR-10 models trained with $B = 10$ using Precision-weighted Federated Learning show an improvement of 12\% (Figure \ref{fig:iid_mnist_fashion} \emph{(c)}) with more stable predictions. This improved accuracy on CIFAR-10 could indicate that there is greater heterogeneity in models trained on natural images than in models trained on grayscale images, even in an IID setting.

\begin{table*}[ht!]
  \centering
  \caption{Comparison of test-accuracy results  (IID data distributions)}
  \label{tab:iid}
  \begin{tabular}{p{2cm}|cc|cc|cc}
    \toprule
     & \multicolumn{2}{c|}{MNIST} & 
        \multicolumn{2}{c|}{Fashion-MNIST} &
        \multicolumn{2}{c}{CIFAR-10} \\
     & FedAvg & PW & 
       FedAvg & PW & 
       FedAvg & PW \\ 
    \midrule
    B = 10 & 
    $0.99 \pm 0.002$ & 
    $0.99 \pm 0.002$ & 
    $0.93 \pm 0.009$ & 
    $0.93 \pm 0.008$ &  
    $0.69 \pm 0.045$ & 
    $\textbf{0.77} \pm \textbf{0.019}$ \\
    B = 25 & 
    $0.99 \pm 0.002$ & 
    $0.99 \pm 0.002$ & 
    $0.93 \pm 0.010$ & 
    $0.93 \pm 0.010$ & 
    $0.77 \pm 0.004$ &
    $0.77 \pm 0.018$ \\
    B = 50 & 
    $0.99 \pm 0.003$ & 
    $0.99 \pm 0.003$ & 
    $0.93 \pm 0.011$ & 
    $0.93 \pm 0.011$ & 
    $0.76 \pm 0.023$ &
    $0.76 \pm 0.013$ \\
    B = 100 & 
    $0.99 \pm 0.004$ & 
    $0.99 \pm 0.004$ &
    $0.93 \pm 0.013$ & 
    $0.93 \pm 0.012$ &
    $0.76 \pm 0.014$ &
    $0.76 \pm 0.011$\\
    B = 200 & 
    $0.99 \pm 0.006$ & 
    $0.99 \pm 0.006$ &
    $0.93 \pm 0.016$ & 
    $0.93 \pm 0.015$ &
    $0.76 \pm 0.014$ &
    $0.76 \pm 0.011$\\
    \hline
    \multicolumn{7}{r}{\small Averaged results using 1 epoch (MNIST and Fashion-MNIST) and 10 epochs } \\

  \end{tabular}
\end{table*}

\begin{figure}[h]
\includegraphics[width=1\linewidth, height=5cm]{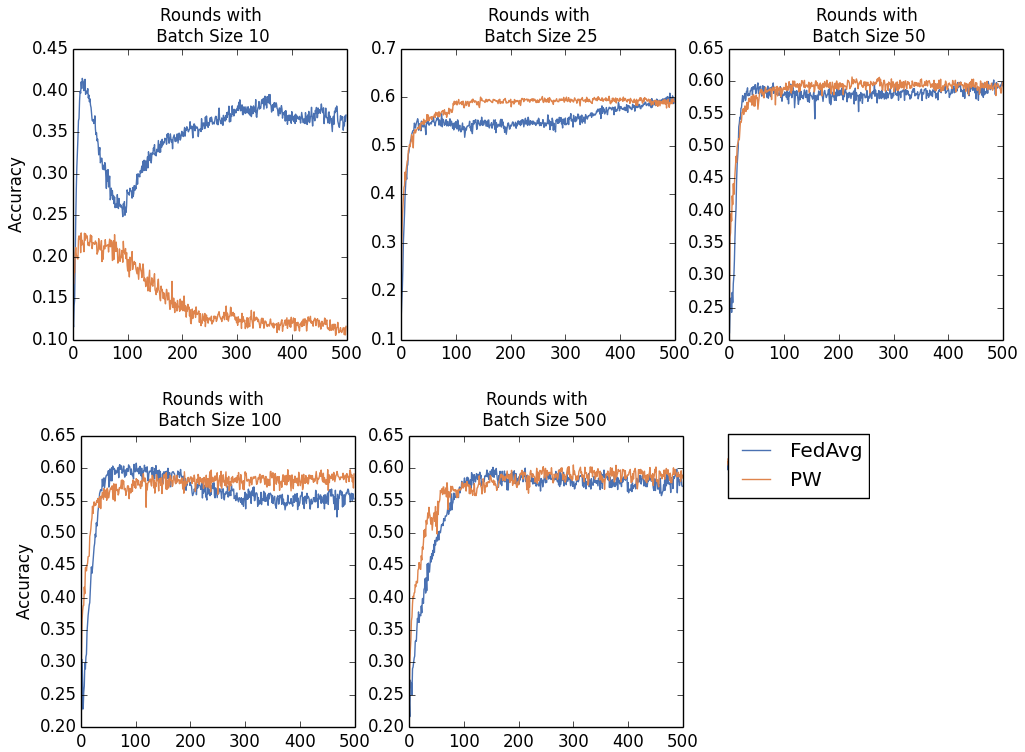}
\caption{CIFAR-10}
\caption{Test-accuracy increases as batch size is larger with non-IID partitions. Aggregation methods:  Federated Averaging (FedAvg) and Precision-weighted Federated Learning (PW). }
\Description{CIFAR-10 test-accuracy scores using non-IID partitions}
\label{fig:noniid}
\end{figure}

As discussed in the introduction section, we hypothesized improvements on the performance of models whose training data is highly-heterogeneous in nature. The comparison results of performance using Non-IID data partitions are given in Table ~\ref{tab:non-iid}. As we observe, both methods perform poorly with a batch number of $B=10$ and more notably in Precision-weighted Federated Learning, which is more sensitive to the noise present in the input images. This behavior of Federated Averaging is comparable with other related work in Federated Learning \cite{zhao2018federated} and its effects are also visible in Precision-weighted Federated Learning. However, with larger batch sizes, higher test-accuracy and more stable predictions are obtained, starting from the first round irregardless of the dataset (Figure \ref{fig:noniid}) This indicates that the estimations of variance are effectively used to computed a weighted average, resulting in more effective penalization of the model's uncertainty at the client level. For MNIST, our method can obtain increases in test-accuracy of up to of 9\% with $B = 200$. The results of Fashion-MNIST show the highest increment of 18\% in the test-accuracy with $B = 200$. Similarly, the highest accuracy of CIFAR-10 improves by 5\% with $B= 100$. These results demonstrate that our first hypothesis is confirmed only when models are trained with a batch size of $B = 25$ or higher.
\begin{table*}[t!]
  \caption{Comparison of test-accuracy results  (non-IID data distributions)}
  \label{tab:non-iid}
  \begin{tabular}{p{2cm}|cc|cc|cc}
    \toprule
     & \multicolumn{2}{c|}{MNIST} & 
        \multicolumn{2}{c|}{Fashion-MNIST} &
        \multicolumn{2}{c}{CIFAR-10} \\
     & FedAvg & PW & 
       FedAvg & PW & 
       FedAvg & PW \\ 
    \midrule
    B = 10 &   
    $0.98 \pm 0.026$ & 
    $0.98 \pm 0.014$ & 
    $0.85 \pm 0.028$ & 
    $0.82 \pm 0.031$ & 
    $0.34 \pm 0.052$ & 
    $0.16 \pm 0.052$ \\
    B = 25 & 
    $0.97 \pm 0.029$ & 
    $\textbf{0.98} \pm \textbf{0.028}$ & 
    $0.85 \pm 0.048$ & 
    $0.85 \pm 0.024$ & 
    $0.51 \pm 0.053$ &
    $\textbf{0.53} \pm \textbf{0.054}$ \\
    B = 50 & 
    $0.97 \pm 0.053$ & 
    $\textbf{0.98} \pm \textbf{0.035}$ & 
    $0.79 \pm 0.048$ & 
    $\textbf{0.86} \pm \textbf{0.035}$ & 
    $0.58 \pm 0.048$ &
    $\textbf{0.60} \pm \textbf{0.027}$ \\
    B = 100 & 
    $0.95 \pm 0.071$ & 
    $\textbf{0.98} \pm \textbf{0.055}$ &
    $0.77 \pm 0.046$ & 
    $\textbf{0.86} \pm \textbf{0.040}$ &
    $0.56 \pm 0.059$ &
    $\textbf{0.59} \pm \textbf{0.041}$\\
    B = 200 & 
    $0.90 \pm 0.083$ & 
    $\textbf{0.98} \pm \textbf{0.058}$ &
    $0.73 \pm 0.052$ & 
    $\textbf{0.86} \pm \textbf{0.050}$ &
    $0.59 \pm 0.07$ &
    $0.59 \pm 0.049$ \\
    \hline
    \multicolumn{7}{r}{\small Averaged results using 1 epoch (MNIST and Fashion-MNIST) and 10 epochs (CIFAR-10) } \\ 

  \end{tabular}
\end{table*}

\subsection{Reliability}
 
The reliability index is an important element to consider in the evaluations of the performance of machine learning systems. In this study, we compute the reliability index defined in \cite{maniruzzaman2018accurate} as the ratio of the standard deviation of the test-accuracy and mean value of the test-accuracy accuracy as shown in Equation \ref{eq:reliability-index}. 

\begin{eqnarray}
\xi_k(\%) = \left( 1- \frac{\sigma_n}{\mu_n} \right) x 100
\label{eq:reliability-index}
\end{eqnarray}

\noindent, where $\sigma_n$ is the standard deviation and $\mu_n$ is the mean of test-accuracy scores per batch. Consequently, the overall system reliability index can be computed by averaging all of the reliability indexes as expressed in Equation \ref{eq:overall-reliability-index}. Table \ref{tab:reliability-index} quantifies the computed reliability index per batch size and shows the overall system stability, which confirms that Precision-weighted Federated Learning reaches optimal performance, except for CIFAR-10 in a non-IID. This is due to the sensitivity of our method with small batch sizes, compromising performance.    

\begin{eqnarray}
\xi(\%) = \left( \frac{\sum_{k=1}^K\xi_k}{N} \right)
\label{eq:overall-reliability-index}
\end{eqnarray}

\begin{table*}[ht]
  \centering
  \caption{Reliability index across batches (IID data distributions)}
  \label{tab:reliability-index}
  \begin{tabular}{p{2cm}|cc|cc|cc}
    \toprule
     & \multicolumn{2}{c|}{MNIST} & 
        \multicolumn{2}{c|}{Fashion-MNIST} &
        \multicolumn{2}{c}{CIFAR-10} \\
     & FedAvg & PW & 
       FedAvg & PW & 
       FedAvg & PW \\ 
    \midrule
    B = 10 & 
    $99.80$ & 
    $99.80$ & 
    $99.03$ & 
    $99.14$ & 
    $93.47$ & 
    $97.52$   \\
    B = 25 & 
    $99.80$ & 
    $99.80$ & 
    $98.92$ & 
    $98.92$ & 
    $94.83$ & 
    $97.67$  \\
    B = 50 & 
    $99.70$ & 
    $99.70$ & 
    $98.81$ & 
    $98.81$ & 
    $96.99$ & 
    $98.29$  \\
    B = 100 & 
    $99.60$ & 
    $99.60$ & 
    $98.60$ & 
    $98.70$ & 
    $98.17$ & 
    $98.55$  \\
    B = 200 & 
    $99.40$ & 
    $99.40$ & 
    $98.27$ & 
    $98.38$ & 
    $98.29$ & 
    $98.55$  \\
    \hline
      & 
    $99.66$ & 
    $99.66$ & 
    $98.73$ & 
    $\textbf{98.79}$ & 
    $96.35$ & 
    $\textbf{98.11}$  \\
  \end{tabular}
\end{table*}

\begin{table*}[ht]
  \centering
  \caption{Reliability index across batches (non-IID data distributions)}
  \label{tab:reliability-index}
  \begin{tabular}{p{2cm}|cc|cc|cc}
    \toprule
     & \multicolumn{2}{c|}{MNIST} & 
        \multicolumn{2}{c|}{Fashion-MNIST} &
        \multicolumn{2}{c}{CIFAR-10} \\
     & FedAvg & PW & 
       FedAvg & PW & 
       FedAvg & PW \\ 
    \midrule
    B = 10 & 
    $97.34$ & 
    $98.57$ & 
    $96.71$ & 
    $96.22$ & 
    $84.62$ & 
    $67.70$   \\
    B = 25 & 
    $97.02$ & 
    $97.13$ & 
    $94.34$ & 
    $97.17$ & 
    $89.57$ & 
    $89.87$  \\
    B = 50 & 
    $94.54$ & 
    $96.42$ & 
    $93.95$ & 
    $95.94$ & 
    $91.72$ & 
    $95.49$  \\
    B = 100 & 
    $92.56$ & 
    $94.37$ & 
    $94.00$ & 
    $95.37$ & 
    $89.52$ & 
    $93.09$  \\
    B = 200 & 
    $90.75$ & 
    $94.06$ & 
    $92.89$ & 
    $94.16$ & 
    $88.03$ & 
    $91.75$  \\
    \hline
      & 
    $94.44$ & 
    $\textbf{96.11}$ & 
    $94.38$ & 
    $\textbf{95.77}$ & 
    $88.69$ & 
    $87.58$  \\
  \end{tabular}
\end{table*}

\subsection{Increasing Participating Clients}

Inspired by McMahan \emph{et al.'s} original paper \cite{mcmahan2016communication}, we experiment with the client fraction $C$ that controls the amount of multi-client parallelism. To this regard, we investigate the number of communication rounds necessary to achieve target test-accuracy of 75\%, 80\%, and 85\% for models trained with Fashion-MNIST. For this purpose, the predictive models used a fixed batch size $B = 100$ and epoch $E = 1$. The training data was split into 100 participants and evaluated speed for every 10, 20, 50, and 100 clients participating in the aggregation in parallel.  

Table \ref{tab:client-speedup} provides the number of communication rounds needed to reach the aforementioned test-accuracy scores as well as their corresponding speedup. We observe a negative correlation indicating that an increase in participants reduces the number of communication rounds irregardless of the number of participants. This behavior is in alignment with McMahan \emph{et al.'s} work in  \cite{mcmahan2016communication}. Given this setup, Precision-weighted Federated Learning misses the first target with 10 and 50 clients, but it can reach subsequent target score up to $20x$ faster with 10 clients and $37x$ with 100 clients participating concurrently. Thus, we see that with a small client fraction ($C = 0.1$; that is 10 client per round), a good balance between computational efficiency and convergence rate can be obtained.

\begin{table*}[t]
  \caption{Number of rounds and speedup relative to Federated Averaging to reach different test-accuracy values on Fashion-MNIST.)}
  \label{tab:client-speedup}
  \begin{tabular}{|l|c c|c c|c c|c c|}
    \toprule
      & \multicolumn{2}{c|}{C = 0.1} &
      \multicolumn{2}{c|}{C = 0.2} & 
      \multicolumn{2}{c|}{C = 0.5} &
      \multicolumn{2}{c|}{C = 1.0} \\ 
     ACC & FedAvg & PW & 
     FedAvg & PW &
     FedAvg & PW & 
     FedAvg & PW\\ 
    \midrule
    75\% & 47 & 50 &
    65 & 35 (19x) &
    21 & 23 &
    17 & 12 (14x)\\
    
    80\% & 149 & 125 (12x) &
    134 & 66  (20x) &
    153 & 57 (27x) &
    44 & 27 (16x)\\
   
    85\% & 641 & 319 (20x) &
    671 & 225 (30x) &
    473 & 279 (17x) &
    286 & 78 (37x)\\
   
  \bottomrule
\end{tabular}
\end{table*}

\subsection{Variance Analysis}

In this paper we demonstrated that combining widely disparate sources can hide important features useful for discrimination, leading to limitations in the collaborative learning experience. Owing to this, Precision-weighted Federated Learning considers the inverse of the estimated variance to compute a weighted average. Given Equation \ref{eq:pw}, this algorithm operates under the assumption that weights with large variance estimations across sources reduces the quality of the analysis and therefore should have a smaller impact in the aggregation.                 

\begin{figure}[h]
\begin{subfigure}{0.4\textwidth}
\includegraphics[width=1.1\linewidth, height=4cm]{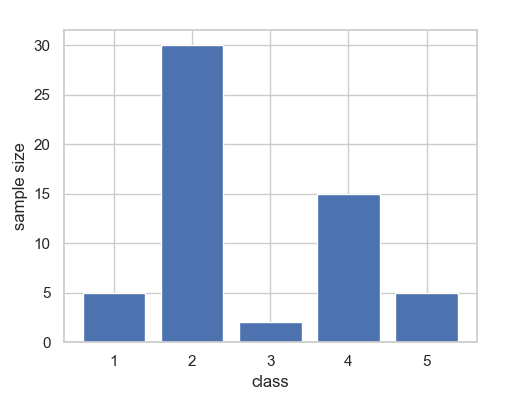} 
\caption{Client 1}
\label{fig:subim1}
\end{subfigure}
\vspace{0.5cm}
\begin{subfigure}{0.4\textwidth}

\includegraphics[width=1.1\linewidth, height=4cm]{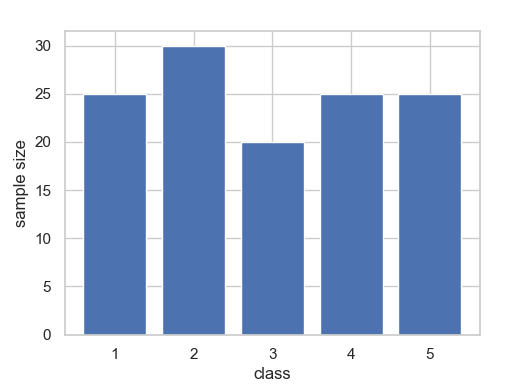}
\caption{Clients 2, 3, 4}
\label{fig:subim2}
\end{subfigure}
\caption{Class distribution per client. One client using an non-IID unbalanced partition, others using an IID partition. }
\Description{Data distribution}
\label{fig:unbalanced-experiment}
\end{figure}

To explain the effects of variance in the generalization of the global model using Precision-weighted Federated Learning, we trained 4 clients with a fixed batch size $B = 10$ and epoch $E = 1$ for 100 communication rounds. The training data of CIFAR-10 was distributed among three clients with IID partitions and a single client with a non-IID partition (Client 1). In this regard, three clients receive a large number of training samples per class, whereas one of the them receives a considerably small number of training samples (Figure \ref{fig:unbalanced-experiment}). This is to maximize the expression of variation across clients. 

\begin{figure}[h]
  \centering
  \includegraphics[width=70mm]{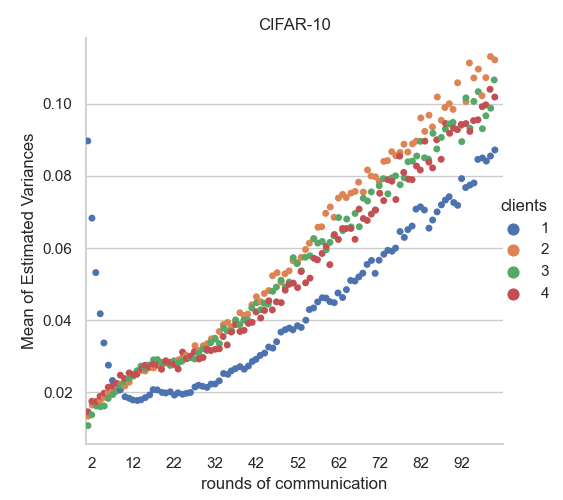}
  \caption{Effect of variance in the generalization of the global model. Each data point represents the mean of the inverse variances per client at a given communication round. Data points in the "Mean of Estimated Variance" graph were normalized between 0 and 1.}
  \Description{variance and model generalization}
  
  \label{fig:variance}
\end{figure}

After model training and before the aggregation, we average the inverse of the estimated variance of the stochastic gradient per client before it is aggregated and plot it. Figure \ref{fig:variance}. Given a small number of training samples, the amount of intra-variability computed for Client 1 is significantly smaller than other models. Consequently, the inverse of the variance for this client is high and therefore the penalization of weights is greater. This behavior is evident since the beginning of the learning cycle and causes a reduction of the inverse of the variance as training continues.

\begin{figure}[h]
  \centering
  \includegraphics[width=70mm]{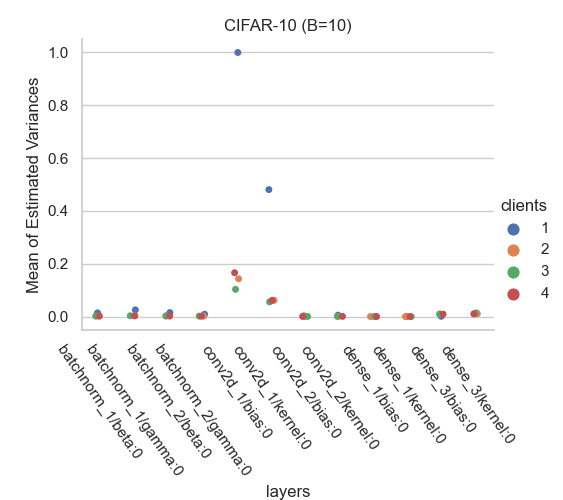}
  \caption{Category plots showing the dispersion of clients per layer at the first round. Data points in the "Mean of Estimated Variance" graph were normalized between 0 and 1.}
  \Description{box plot round one}
  \label{fig:cat-plot-layers}
\end{figure}

Given a small number of training samples, the amount of intra-variability computed for Client 1 is significantly smaller than other models. Consequently, the inverse of these variances for this client is high and therefore the penalization of weights is greater (Equation \ref{eq:pw}). This behavior is evident since the first communication round and results in a better model generalization if achieved. Alternatively, models with larger training samples provide weights with higher quality and their penalization is minimum. Figure \ref{fig:cat-plot-layers} shows the inverse of the estimated variance per weight and client. With this view we can identify \textit{conv2d/bias} and \textit{conv2d/kernel} with the highest mean of the inverse variance. This suggests that the Adam optimizer could not capture the most prominent characteristics that make up the training data, for these layers, due to the limited number of training passes.

\section{DISCUSSION}

Federated Learning is a promising solution to the analysis of privacy-sensitive data distributed globally across clients. At the core of Federated Learning is Federated Averaging, an aggregation algorithm that consolidates the weighted average of  distributed machine learning models into a global model shared with every client participating in the learning cycle. In this paper, we hypothesized that Federated Averaging underestimates the full extent of heterogeneity of data across participants, leading to a reduction in the statistical power and quality of predictions, and thus proposed Precision-weighted Federated Learning. Our method averages the weights of individual sources by the inverse of the estimated variance. When weighting machine learning models differently, it must be noted that different aggregation algorithms may yield different results under different circumstances. Our method shows the greatest advantages when the data is highly-heterogeneous across clients.

Our first hypothesis postulates that not accounting for variation across clients may lead to a reduction of statistical power when combining data form multiple sources. We confirmed this hypothesis by showing that models trained with batch size $B >= 25$ and Precision-weighted Federated Learning can obtain a $2\%$ improvement with MNIST, $14\%$ with Fashion-MNIST, and $9\%$ with CIFAR-10 using non-IID partitions Nevertheless, the presented algorithm can still be improved. With a batch size $B = 10$, our method is sensitive to the noise introduce by individual sources, degrading the performance of the method. These results prove the limits of our algorithm. Alternatively, when we compare our method with those models trained with IID partitions, our method shows comparable results to those of Federated Averaging. This suggest that the inter-variance estimations were small due to the large number of training samples and uniform distribution of classes among participants, leading to more confident predictions.

Our second hypothesis addresses convergence speed and supports the idea that the use of estimated variance can capture better representation of intricate features dispersed across sources, resulting in an acceleration of the learning process. We confirmed this hypothesis by demonstrating that our method can reach test-accuracy targets faster, and with less communication rounds between targets, than Federated Averaging. With Fashion-MNIST, we also obtained a $24x$ speedup (with only 10 clients trained in parallel) than Federated Learning. This suggest that our method reduces the communication costs required between rounds. Although it is possible to achieve higher test-accuracy by using more complex state-of-the-art architectures, our goal in this study was to explore the statistical challenges, especially when the training data is non-IID. Therefore, we measure the performance of both aggregation method with simple network architectures.

Although the aggregation of model parameters, rather than raw individual client data, represents a significant step towards privacy preservation, the Precision-weighted Federated Averaging algorithm remains vulnerable to inference attacks, as the model parameters still contain information about data. This is a limitation of the general Federated Learning protocol and is not exclusive to our approach. Recently, Geyer \emph{et al.} \cite{geyer2017differentiallyprivate} and Truex \emph{et al.} \cite{truex2018hybrid} introduced frameworks that preserve client-level differential privacy. However, Melis \emph{et al.} demonstrated that privacy guarantees at the client-level are achieved at the expense of model performance and are only effective when the number of clients participating in the aggregation is significantly large, thousands or more \cite{melis2019exploiting}. Owing to this, we will examine the behavior and performance of the Precision-weighted Federated Learning scheme combined with Differential Private Federated Learning \cite{dwork2014algorithmic, wei2020federated} as a future work.

\section{CONCLUSION}

In this paper we presented an novel aggregation algorithm for computing the weighted average of distributed DNN models trained in a Federated Learning environment. It does not require sharing raw private data. Instead, this algorithm takes into consideration the second raw moment (uncentered variance) of the stochastic gradient estimated from the Adam optimizer to compute the weighted average of distributed machine learning models. Precision-weighted Federated Learning was benchmarked with MNIST, Fashion-MNIST and CIFAR using two data distribution strategies (IID and non-IID). When compared to Federated Averaging, this algorithm was shown to provide significant advantages when the data is highly-heterogeneous across clients, and showed comparable test-accuracy when the data is uniformly distributed across clients. Demonstrating that including the variability across models in the aggregation results in a more effective and faster option for averaging distributed machine learning models having complex data with a large diversity of features in its composition. With these advantages, Precision-weighted Federated Learning show promise in comprehensive exploratory analyses of sensitive biomedical data distributed across medical centers. Thus, in future work we will examine the feasibility of this method in medical image classification tasks.

\bibliographystyle{ACM-Reference-Format}
\bibliography{sample-base}

\end{document}